# Aesthetics of Neural Network Art

Aaron Hertzmann

Adobe Research

**Abstract**

This paper proposes a way to understand neural network artworks as juxtapositions of natural image cues. It is hypothesized that images with unusual combinations of realistic visual cues are *interesting,* and, neural models trained to model natural images are well-suited to creating interesting images. Art using neural models produces new images similar to those of natural images, but with weird and intriguing variations. This analysis is applied to neural art based on Generative Adversarial Networks, image stylization, DeepDream, and Perception Engines.

*"I am trying to find interestingness which is a search that never ends since it is in its nature to melt away like a snowflake in your hand once you have captured it."* — *Mario Klingemann* [1]

## 1. Introduction

In the past few years, many artists have begun to explore neural networks as artistic tools, and their works have begun to appear in cutting-edge "Artificial Intelligence" art shows as well. Notable artists working in this space include Memo Akten, Refik Anadol, Mario Klingemann, Trevor Paglen, Anna Ridler, Helena Sarin, and Tom White. Some of these artists, including Salavon and Paglen, operate within the mainstream art world of major galleries, museums, and auctions, whereas others, like Robbie Barratt and Helena Sarin, have gained prominence primarily due to their inspiring Twitter feeds of neural network art.

In computer vision and perceptual psychology, image perception is often analyzed in terms of visual *cues*. In principle, any image property can be a cue, including color, texture, local parts, overall shape, as well as learned features. In real images, each cue provides complementary information to the viewer.

This essay hypothesizes that images become visually *interesting* when they comprise *unusual juxtapositions of realistic visual cues*. Moreover, neural network art is interesting because it exploits networks designed to model the cues of real-world images. By tweaking these natural image models, artists produce images that obey some natural image cues but not others, and thus are visually intriguing, evocative, weird, and surreal.

In other words, modern neural models lend themselves to creating interesting imagery, because they were designed modern real images, and so modifying them creates realistic but unreal images. Models trained on photographs often appear to be a sort of photography of imaginary things.

This simple description encapsulates many different kinds of artworks, and provides a recipe for new kinds of visual art in the future.

These modern works have fascinating modern conceptual components as well. Much of their interest comes from their relation to the algorithms and "AI" software that plays an increasing role in our lives. However, this essay is solely concerned with the visual qualities of art, primarily interestingness, as distinct from attractiveness or other properties.

## 2. Juxtapositions of Cues: A Theory of Image Interestingness

The first part of this paper's theory is that images that juxtapose unusual sets of realistic cues are *interesting*.

In order to understand this theory, consider how robust human vision is. Each of us is capable of recognizing and understanding scenes based on many different visual cues: we can recognize objects just from their outlines or from grayscale images; we can recognize blurry objects just from their colors; we can recognize a person walking just from their gait. Many of these cues are statistical, such as an object's average color. Most of the time, these cues produce a consistent understanding of the world: if we see an object that is red and apple-shaped and has apple texture, it is easy to recognize it as an apple.

Interesting images arise when these cues conflict. As a simple example, a blue object that is apple-shaped and has apple texture (Figure 1) gives conflicting cues about what it is, and demands an explanation. Different combinations of cues will give more or less surprising images. By "Interesting", we mean that an image draws one's attention and entices the viewer to study it further. This theory builds on an earlier one I proposed [2]: interesting images are those that our visual system perceives with conflicting interpretations.

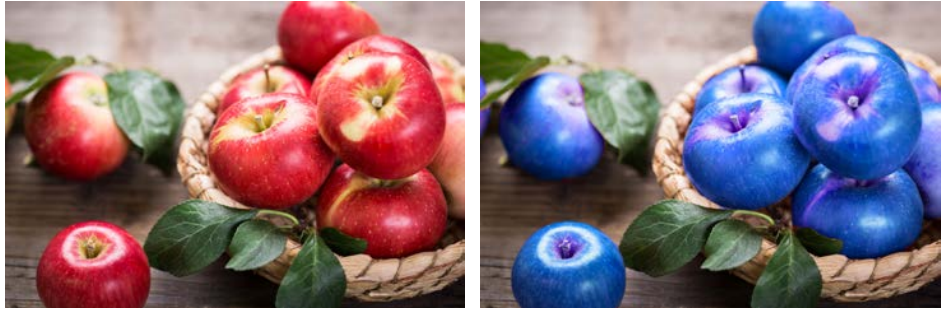

**Figure 1.** Red Delicious, Blue Delicious. The image on the right is more interesting, because the shape and texture give a strong impression of apples, but the colors conflict with this.

Interestingness is contextual, and fades as it loses novelty; an image of a blue apple is not all that odd these days. When Andy Warhol first produced photographic portraits of Marilyn Monroe and others with unrealistic color palettes, such combinations must have seemed much more surprising. Image morphing is another example: it adjusts shape but not color or texture, and was once a novel and intriguing digital effect.

Conceptual juxtapositions abound throughout art and pop culture, like cats in business suits, Dalí's melting clocks, or Banksy's protestor throwing flowers. These, too, can be interesting at first, because they demand explanation; however, this essay focuses on novel low-level visual cues.

In the rest of this paper, I explore this theory in the context of several kinds of neural network art, as an explanation for why this technology is so well-suited to creating visually interesting art.

**3. Why GAN images are interesting**

The most prominent tool in neural art at the moment is the Generative Adversarial Network (GAN) [3]. Given a large collection of images of a specific class (such as faces or landscapes), a GAN is trained to produce new images that look like they also came from that class. However, GANs operate in terms of image cues that are difficult to explain; they are not just manipulating simple properties like color and texture.

GANs are the latest in a long line of research in natural image modeling. Vision neuroscience has long sought to build statistical models of "natural images," originally referring to images in prehistorical environments, but now referring to any realistic photos of the world. This work has led to insight on the human vision system [4] [5]. Over the years, while developing these models, researchers sampled images from these models; as the models got more realistic, so did the sampled images (Figure 2). These algorithms eventually led to texture synthesis and style transfer algorithms [6] [7].

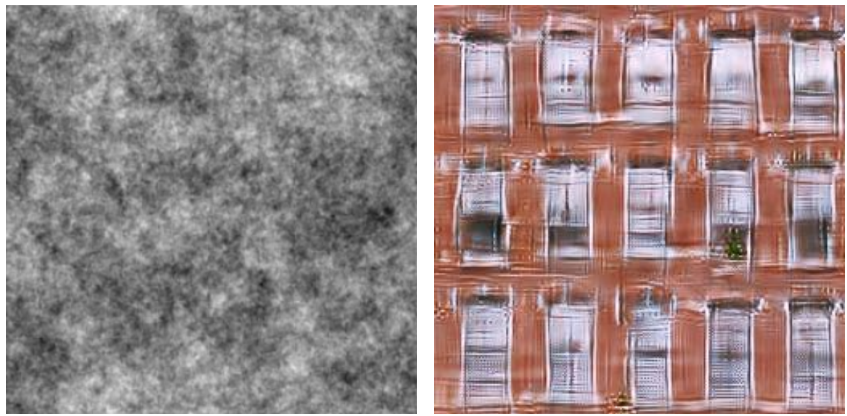

Figure 2. Image sampled from early image models: a linear Fourier basis model [8], and matching gradient filter statistics of a source photograph (not shown) [9]

A GAN model comprises a Generator and a Discriminator, trained jointly. The Generator is used to sample new images. The Generator is optimized according to the Discriminator, which is a classification network meant to recognize whether images are "real." Hence, the Generator is optimized according to a form of natural image cues.

More specifically, the trained Generator has two components: a sampler that randomly generates a low-dimensional "latent vector," often denoted **z**, and a neural network that generates an image from **z**. The vector **z** can be thought of as the "genetic code" of an image: the neural network creates an image by following the "instructions" given in **z** to create an image. As described above, the Generator has been trained to produce images with some of the statistics of natural images.

What does the Generator network actually do? This is an open question; however, following from some studies on discriminative networks [10] [11] and GANs [12], we hypothesize the following: GANs *construct* images, first generating arrangements of scene labels, filling in object labels from the scene labels, and then placing detail texture according to these sets of labels. In other words, GAN generator nets have learned how to plausibly compose objects together in scenes and to texture the object parts, following the instructions encoded in the vector **z**. They do so, trying to create images with as realistic cues as possible.

When new **z** vectors are randomly sampled, they often produce many highly-realistic images from scratch: images indistinguishable from photographs, but yet totally imaginary. Sometimes these random images do not look entirely real.

At the moment, the most impressive GAN model is BigGAN [13]. This method was trained on ImageNet, an enormous and diverse collection of real-world images; based on the training details in the paper, it has been estimated that model training used an amount of power consumption equivalent to a typical US household's usage over 6 months [14]. Now that it has been trained and released online, anyone can easily download and experiment with it.

Things get really unpredictable when we start to feed bogus "instructions" to the model. This can be seen at the Ganbreeder website, which provides an easy facility for collaborative image remixing by modifying and combining latent **z** vectors (Figure 3). This applies a form of the evolutionary art ideas of Karl Sims [15] and others to BigGAN. The reader is encouraged to spend a few minutes on this site to see some of the images created through this process, and to spend some time trying to understand what makes these images so intriguing and evocative. The images look so real and so diverse in appearance, and yet not real. Our brains recognize them as having the overall elements of real photographs, but cannot recognize them as anything real.

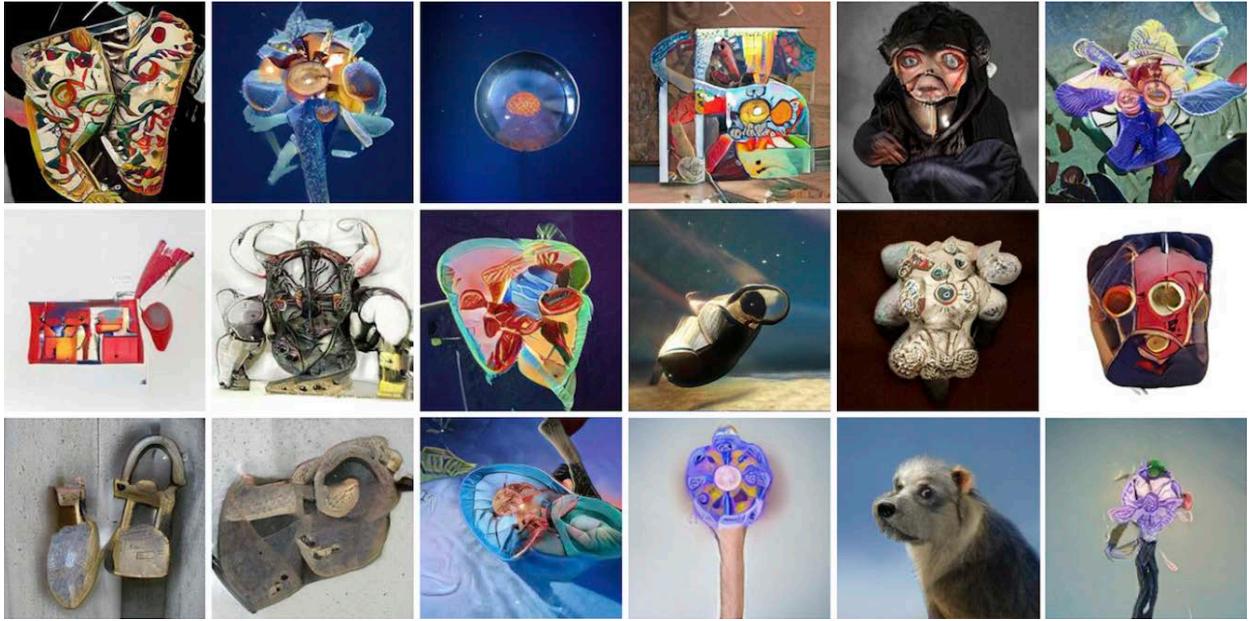

Figure 3: BigGAN images from the GANbreeder website.

Much of the work that artists do with GANs is to explore the latent space and experiment with different ways to generate **z** vectors. Discovering the weird structures that emerge from a model can be a pleasure in itself. For example, Mario Klingemann has tweeted many different ways to explore the BigGAN latent space to find codes that generate interesting, provocative, and beautiful images, e.g., [16]. He and Robbie Barrat have each generated many intriguing animations simply by following continuous paths in **z** space, as was also done in some of the earlier GAN research [17]. In some cases, the GAN is trained on a specific class of images, such as classical portraits in Klingemann's "Memories of Passersby I" installation.

## 4. Image Stylization

We now apply the theory to image stylization algorithms. These algorithms take a natural image as input and output a stylized version, such as a painting. They often express this task as an optimization that trades-off two goals: producing an image that has the overall appearance and structure of the given photograph, while also constructing the image from style elements, such as paint strokes or tiles. This optimization has been expressed both as an explicit optimization of style elements (Figure 4) [6] [18], and as an optimization employing neural image statistics [7]. In each case, the duality between low-level image cues (e.g., brush strokes) and the underlying scene provides a visual tension. That is, the image appears to be a real scene but also appears to be composed of unrelated texture. These conflicting cues provoke interest.

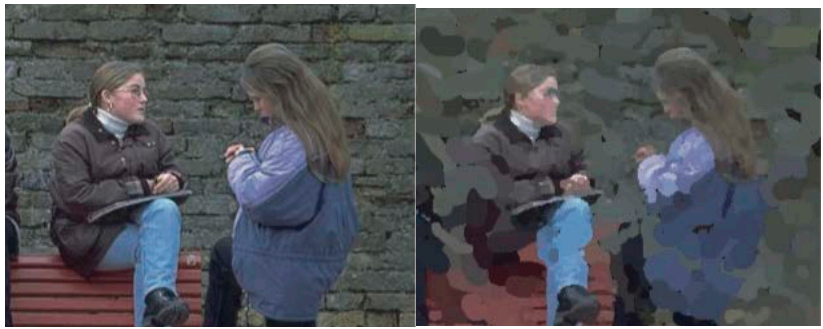

Figure 4: Image stylization expressed as optimizing a limited number of brush strokes to approximate an input photograph [19].

More recently, the DeepDream method [20], which was adopted by artists like Mike Tyka and Gene Kogan, began as an attempt to visualize neural activations. It can be described as trading-off fidelity to an original image against maximizing certain neural activations. It produces images that preserve the overall scene structure but replaces details with unexpected object parts.

**Neural image translation**. Neural image translation methods, like pix2pix [21] and CycleGAN [22], learn to transform images, given before-and-after pairs as training. Rather than attemping to model natural images, they model image *transformations*. Helena Sarin has worked extensively with these methods. By training CycleGAN with her own data, she creates new visual styles from what the model learns from her data. In this case, the GAN creates a new visual style for input imagery, rather than creating new imagery from scratch. Sarin's imagery has a unique blend of hand-made art with neural "artifacts" that suggest entirely new visual styles. They create still lifes and landscape drawings with the appearance of unique and novel drawing styles that, while similar to real drawing styles, nonetheless seem like something no human would have come up with on their own (Figure 5).

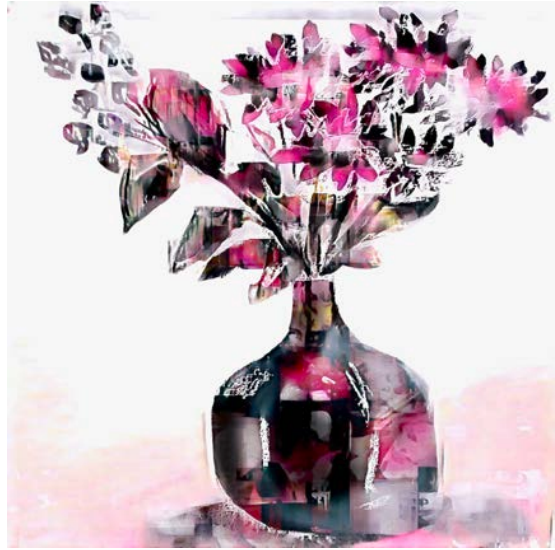

Figure 5. GAN-stylized image by Helena Sarin [23]

**5. Image Statistics and Abstract Expressionism**

Another class of works also arose from attempts to visualize neural network classifiers. For example, Nguyen et al. [24] evolved procedural images to maximize object recognition scores; these highly-abstracted versions of object classes were later shown in art galleries under the name "Innovation Engines" [25]. Perhaps unexpectedly, these works suggest a way to interpret abstract expressionism.

**Perception Engines.** Tom White's "Perception Engines" series [26] combines classifier visualization with stroke-based optimization [18], while producing an intriguing new take on abstract art. In this work, he optimizes stroke layouts according to image classifiers, i.e., producing images that trained classifiers recognize as specific objects. These images are barely-recognizable abstractions of real-world objects, but they still evoke similar responses.

These images are optimized according to natural image classifiers that are trained to distinguish photographs from one another, for example, looking for cues that distinguish photos of, say, violins from any other object class. They do not attempt to distinguish photographs from non-photographs. Hence, optimizing strokes by these classifiers produces certain discriminative cues without attempting to make the images realistic.

In each of these cases, there is a tension between the apparent abstraction of the image strokes and the object class cues specified by the classifier. This is most apparent in White's "Pitch Dream", a recent series of stroke-based images that trigger current "adult content" filters (Figure 6) [27], demonstrating how inaccurate they can be. These images are pure abstraction, yet, they feel oddly pornographic. Given ongoing battles in our society over which kinds of images are offensive and why, they provide an interesting example of how something can seem somehow "obscene" while being purely abstract.

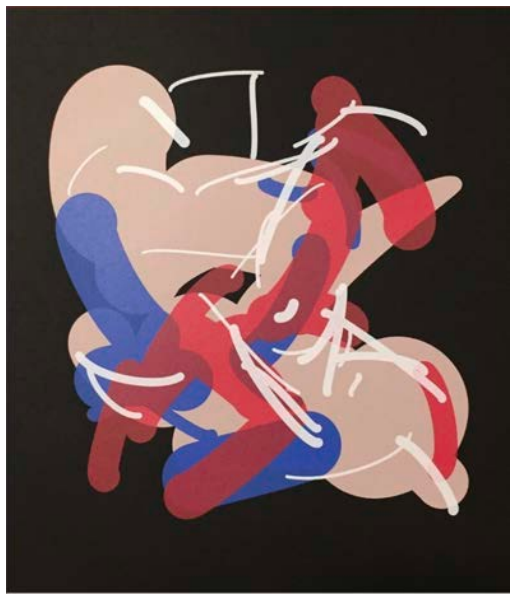

Figure 6. "Pitch Dream," by Tom White: Strokes optimized by Tom White to maximize pornography filter responses [27]

**Abstract Expressionism and Natural Images.** White's work provides an example for how art can evoke responses tied to real world scenes while being purely abstract. More generally, it provides a possible model for more conventional abstract expressionism. As described above, realistic drawings get some of their appeal from depicting natural scenes with specific style elements. Even though one might think of landscape painting and abstract expressionism as total opposites—landscape painting is often trite and conventional, and abstract expressionism was, at least for some, the very definition of unapproachable modernism—they both may exploit a similar contrast between natural image cues and abstract compositional elements.

In particular, Taylor et al. [28] showed that Jackson Pollock's drip paintings obey a particular form of normal image statistics. Following this observation, a number of authors discovered commonalities in the image statistics of many other artists [29]. This suggests the following interpretation of Pollock's paintings: they are interesting because they juxtapose some overall cues of natural scenes with fine details that are completely non-naturalistic. They capture the overall sense of real scenes with none of the specific content.

In Richard Diebenkorn's work, the connection between abstract expressionism and landscape painting is more explicit: his landscape works seem to fall on a continuum from slight abstraction (*Cityscape #1*), to heavy abstraction (*Berkeley #44*), to total abstraction (*Ocean Park* series). His entirely-abstract *Ocean Park* series explicitly references landscapes in the title, the colors, and the layout of abstract shapes. His images have the pleasing feeling of a sunny ocean landscape, but the novel abstractions of modernism. Piet Mondrian too, began as a painter of natural scenes, and his gradual movement toward abstraction shows in his purely abstract works as well.

In this way, Tom White's abstract stroke arrangements can be thought of as an implementation of an explicit model of some abstract expressionism: creating abstract arrangements with some of the statistics of natural images.

## 6. From Interestingness to Expression

This paper presents a sort of recipe for making interesting images: begin with the latest research in natural image modeling, and then modify some elements of the model while preserving others. This will naturally produce unusual juxtapositions of natural cues, which, in turn, will be visually interesting.

As new models are developed, trained, and released in the coming years, we can expect that artists will continue to seize on these models and exploit them to create new and fascinating imagery. Even if new GANs become flawless, artists will still tweak them to produce surprising new imagery.

Interestingness is not beauty, however, and it is not staying power. The DeepDream imagery was briefly fascinating, but, within a year or two, it lost its novelty and appeal. The DeepDream images eventually seemed very repetitive, with the same mutant doggies and eyeballs appearing over and over again. This may be, in part, because the DeepDream tools provided relatively little opportunity for artistic control or artistic expression. The current GANs provide a much richer visual space, leading to considerable enthusiasm [30], but they might also lose their appeal in a few years. Or they might be basis for a new, rich tool for visual expression. Interestingness is a sign that there is something worth investigating further.

These neural images are, in part, the search for new visual languages. Many new kinds of imagery have been discovered. It still remains to use these new visual languages for lasting visual expression. The current level of energy and activity in this space inspires optimism that great things are afoot.

**Acknowledgements.** I am grateful for feedback from Alyosha Efros, Jason Salavon, and Damian Stewart.